\pdfoutput=1
\documentclass[11pt]{article}

\usepackage[]{acl} %review
\usepackage{times}
\usepackage{latexsym}
\usepackage{booktabs}
\usepackage{subcaption}

\usepackage[T1]{fontenc}
\usepackage[utf8]{inputenc}

\usepackage{microtype}
\usepackage{graphicx}

\usepackage{inconsolata}

\title{Leveraging Vision–Language Pre‑training for Human Activity Recognition in Still Images}

\author{Cristina Mahanta, Gagan Bhatia\\
  Department of Computing Science \\
  University of Aberdeen \\
\texttt{\{c.mahanta.24,g.bhatia.24\}@abdn.ac.uk} \\}

\begin{document}
\maketitle
\begin{abstract}
Recognising human activity in a single photo enables indexing, safety and assistive applications, yet lacks motion cues. Using 285 MS‑COCO images labelled walking/running, sitting and standing, scratch CNNs scored 41\% accuracy. Fine‑tuning multimodal CLIP raised this to 76\%, proving contrastive vision‑language pre‑training decisively improves still‑image action recognition in real deployments.
\end{abstract}

\section{Introduction}

Detecting human activities from still images is a challenging problem in computer vision, largely due to the subtle and complex variations inherent in human behaviour. In this work, we address this task using a subset of MSCOCO 2017 validation split introduced by \citet{lin2015microsoftcococommonobjects}, each labeled as walking/running, sitting, or standing. We begin with two baseline models --- Convolutional Neural Network (CNN) and Feedforward Neural Network (FNN) --- and then enhance performance through data augmentations, dropout, weight decay, and early stopping. To utilize broader visual knowledge, we apply transfer learning with pretrained Vision Transformers and contrastive models (e.g., CLIP) and explore multimodal embeddings for richer feature representations. We detail the preprocessing steps, model configurations, and evaluation metrics to enable transparent comparison across all methods.
\begin{figure}[ht]
\includegraphics[width=1\columnwidth]{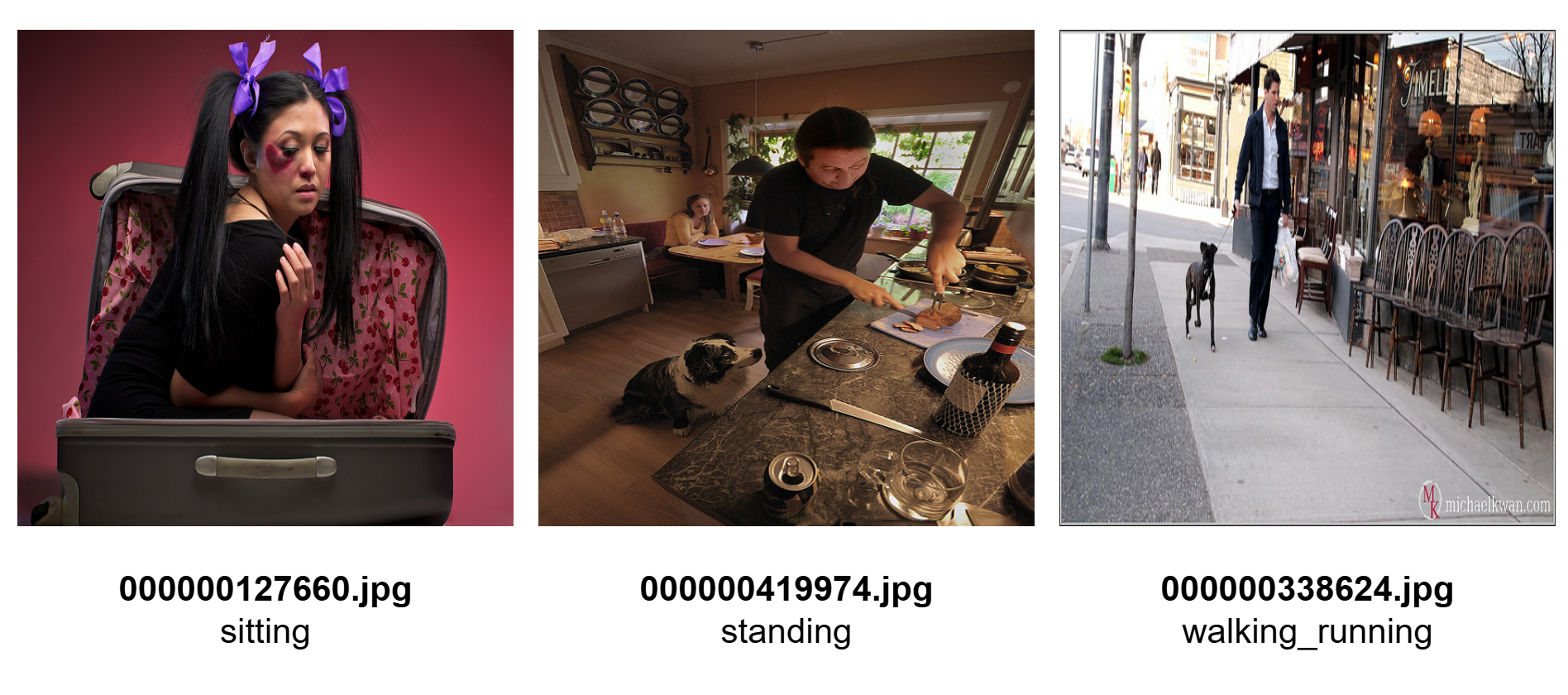}
\centering
\caption{Examples of the training data}
\label{fig:data_preview}
\end{figure}
\section{Description of Data and Methods}

\subsection{Data}
\label{subsec:data}
The dataset for this study is drawn from a carefully curated subset of the Microsoft COCO (Common Objects in Context) validation split, originally introduced by \citet{lin2015microsoftcococommonobjects} and now a gold standard benchmark in computer vision tasks such as object detection, segmentation and image captioning. From the full COCO set, we selected 285 images depicting exactly one of the three human activities—walking/running (98 images), sitting (95 images), or standing (92 images) —yielding a nearly balanced three-way classification problem. All images were downloaded directly via their URLS, and none were discarded due to corruption or missing annotations, confirming complete data integrity. A few examples of the training dataset are shown in Figure \ref{fig:data_preview}.

% \begin{figure}[t]
% \includegraphics[width=1\columnwidth]{figs/plot_scatter.pdf}
% \centering
% \caption{EDA}
% \label{fig:plot_scatter}
% \end{figure}

\begin{figure*}[t]
  \centering
  % ensure each panel is exactly one quarter of textwidth:
  \begin{subfigure}[t]{0.24\textwidth}
    \includegraphics[width=\linewidth]{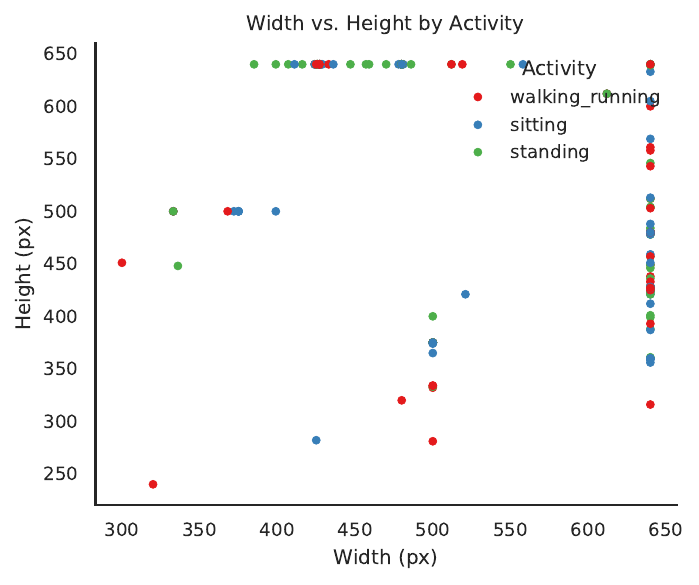}
    \caption{Scatter of width vs.\ height}
    \label{fig:eda_scatter}
  \end{subfigure}\hfill
  \begin{subfigure}[t]{0.24\textwidth}
    \includegraphics[width=\linewidth]{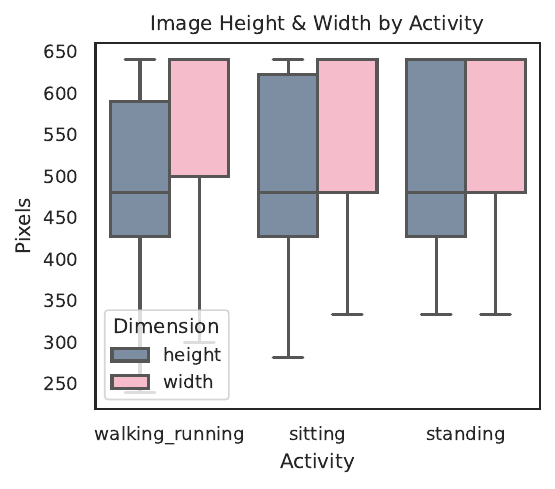}
    \caption{Box‑plots by activity}
    \label{fig:eda_box}
  \end{subfigure}\hfill
  \begin{subfigure}[t]{0.24\textwidth}
    \includegraphics[width=\linewidth]{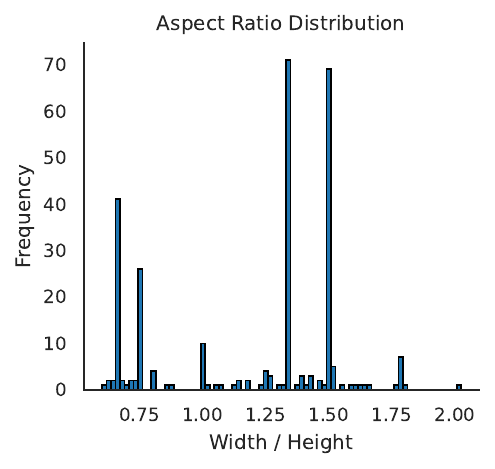}
    \caption{Aspect ratio histogram}
    \label{fig:eda_aspect}
  \end{subfigure}\hfill
  \begin{subfigure}[t]{0.24\textwidth}
    \includegraphics[width=\linewidth]{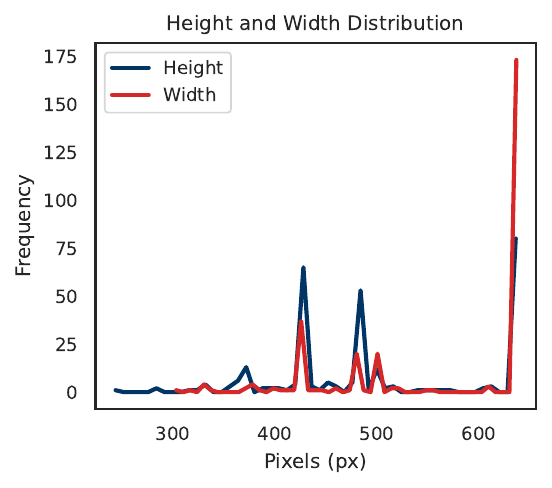}
    \caption{Overlaid height/width distributions}
    \label{fig:eda_overlay}
  \end{subfigure}
   \caption{Exploratory data analysis of image dimensions:  
    (a) width vs.\ height scatter,  
    (b) height/width box‑plots grouped by activity,  
    (c) distribution of aspect ratios, and  
    (d) overlaid height and width histograms.
  }
  \label{fig:eda_all}
\end{figure*}

%Please add the following packages if necessary:
%\usepackage{booktabs, multirow} % for borders and merged ranges
%\usepackage{soul}% for underlines
%\usepackage{xcolor,colortbl} % for cell colors
%\usepackage{changepage,threeparttable} % for wide tables
%If the table is too wide, replace \begin{table}[!htp]...\end{table} with
%\begin{adjustwidth}{-2.5 cm}{-2.5 cm}\centering\begin{threeparttable}[!htb]...\end{threeparttable}\end{adjustwidth}
% \begin{table}[!htp]\centering
% \caption{Generated by Spread-LaTeX}\label{tab: }
% \scriptsize
% \begin{tabular}{lrr}\toprule
% \textbf{label} &\textbf{count} &\textbf{\% of Total} &\textbf{Width}\\\midrule
% \textbf{walking\_running} & 98 & 34.4\%\\
% \textbf{sitting} & 95 & 33.3\%\\
% \textbf{standing} & 92 & 32.3\%\\
% \bottomrule
% \end{tabular}
% \end{table}
\begin{table*}[ht]
  \centering
  \begin{tabular}{lrrrrr}
    \hline
    \textbf{Label}
      & \textbf{Count}
      & \shortstack{\textbf{\% of}\\\textbf{Total}}
      & \shortstack{\textbf{Width}\\\textbf{(mean $\pm$ $\sigma$, px)}}
      & \shortstack{\textbf{Height}\\\textbf{(mean $\pm$ $\sigma$, px)}}
      & \shortstack{\textbf{Aspect Ratio}\\\textbf{(mean $\pm$ $\sigma$)}} \\
    \hline
    \textbf{walking\_running} &  98 &  34.4\% & $576.1 \pm 60.2$  & $492.3 \pm 85.7$  & $1.17 \pm 0.21$ \\
    \textbf{sitting}          &  95 &  33.3\% & $558.2 \pm 68.5$  & $495.8 \pm 95.4$  & $1.19 \pm 0.23$ \\
    \textbf{standing}         &  92 &  32.3\% & $558.7 \pm 64.1$  & $512.2 \pm 87.2$  & $1.22 \pm 0.19$ \\
    \textbf{Overall} & 285 & 100.0\% & $565.7 \pm 99.2$  & $499.4 \pm 100.5$ & $1.20 \pm 0.22$ \\
    \hline
  \end{tabular}

  \medskip
\caption{Dataset and per‐class image statistics. Ranges (width: 300--640\,px; height: 240--640\,px) apply across all classes. No missing or corrupted entries were found.}
  \label{tab:labels_data}
\end{table*}

After performing a detailed exploratory data analysis (EDA), it can be seen that the images range from 300 to 640 pixels in width, clustering around an average of approximately 566 pixels, and from 240 to 640 pixels in height, centred near 499 pixels, as shown in Table~\ref{tab:labels_data}. Moreover, when comparing the distribution plots for walking/running, sitting, and standing, their medians, interquartile ranges, and overall distributions align almost perfectly. This indicates that no activity label consistently contains larger or smaller images. Table~\ref{tab:labels_data} also highlights that the class frequencies remain nearly identical, with each activity accounting for roughly one-third of the dataset, so no label imbalance is expected to bias model training. The scatter and box plots in Figure~\ref{fig:eda_all} further confirm that no class systematically contains larger or smaller images. At the same time, the aspect-ratio histogram exhibits two dominant modes around 1.0 (square image) and 1.33 (4:3), with an overall mean ratio of approximately 1.2. Together, these findings demonstrate that the dataset is inherently balanced and free of resolution- or framing-based biases. This means that any necessary standardisation (e.g. image resizing and normalisation) can be applied uniformly, and the augmentation strategies need to focus only on semantic diversity rather than correcting for class-specific size or aspect ratio artefacts.  

\subsection{Models}
In this work, we compare several neural network architectures and transfer learning approaches for image classification. In the following subsections, we describe each model and the underlying design choices.
\paragraph{CNN and FNN}

The \textbf{CNN\_base} model follows a classic convolutional design: three blocks of $3\times3$ convolutions (padding=1) each succeeded by ReLU activations and $2\times2$ max‐pooling, with the resulting feature map flattened into a two‐layer fully connected classifier. By exploiting local spatial correlations and hierarchical feature extraction, it embodies the core principles of convolutional networks \cite{lecun1998gradient,oshea2015introductionconvolutionalneuralnetworks} and serves as a robust baseline. In contrast, the \textbf{FNN\_base} model treats each image as a flat vector passed through successive dense layers with nonlinearities. Lacking the spatial inductive biases and weight sharing of convolutions, this fully connected architecture consistently underperforms on visual data \cite{Yann2016deep}.
\paragraph{Generalising CNN}
The \textbf{CNN\_gen} model extends the baseline CNN architecture by integrating several regularization and normalization techniques to improve generalization. Batch normalization \cite{ioffe2015batchnormalizationacceleratingdeep} is applied after each convolution to stabilize activations, and dropout layers are interleaved with the convolutional and fully-connected blocks to prevent overfitting. These modifications, combined with data augmentation during training, enable the CNN\_gen model to achieve better performance on unseen data while maintaining the simplicity of convolutional feature extraction.
\paragraph{Transfer Learning for Binary Classification}
We leverage a pretrained Vision Transformer (ViT) backbone, which splits each image into $16\times16$ patches and processes the resulting sequence through standard transformer blocks to capture long‐range dependencies and global context \cite{dosovitskiy2021imageworth16x16words}. Transformers’ self‐attention mechanism and large‐scale pretraining yield highly generalizable feature representations, making them particularly well suited for transfer learning across diverse vision tasks.  In addition, we fine‐tune two modern vision–language encoders --- CLIP \cite{radford2021learning} and SigLIP2 \cite{tschannen2025siglip} --- by replacing their projection heads with a two‐way classifier and optimizing all parameters end‐to‐end under a cross‐entropy objective.  
\paragraph{Transfer Learning for Multiclass Classification}
For multiclass tasks, we adopt the same pretrained ViT, CLIP, and SigLIP2 models, extending each classification head to output $C$ logits, where $C$ is the number of target categories.  All three backbones are fine‐tuned jointly with the new head under cross‐entropy loss, allowing them to adapt their rich, pretrained representations to the specific demands of our domain‐specific multiclass classification problem.  
\paragraph{CLIP Image Embeddings}
We use CLIP embeddings in two complementary ways.  In the first setting, we treat CLIP \cite{radford2021learning} as a pure image encoder: given a image we encode it to obtain a fixed‑length vector of dimension $d$ (typically 512).  These image vectors are paired with their ground‑truth labels to build a simple PyTorch dataset, which we split into training, validation, and test subsets.  We then train a small multilayer perceptron (\texttt{MultimodalClassifier}) on top of the raw CLIP embeddings, consisting of several fully connected layers with ReLU, batch normalization, and dropout, and optimize with cross‑entropy loss.  This setup tests how linearly separable the CLIP image representations are for our target classes.
\paragraph{CLIP Image-Text Embeddings} In the second setting, we take advantage of CLIP’s joint image–text space by also encoding a set of textual label descriptions (e.g., “walking”, “standing”, “sitting”).  We compute the cosine similarity between each image embedding and each label embedding via cosine similarity, producing an $N\times C$ similarity matrix (where $N$ is the number of images and $C$ the number of classes).  Each row of this matrix --- one cosine score per class --- serves as a compact, semantically meaningful feature vector.  We then train a second MLP (\texttt{FeatureClassifier}) on these similarity features, allowing the model to directly leverage the semantic affinity between images and label text without relying solely on the high‑dimensional raw embeddings.  

\subsection{Experimental Approach}
We partitioned the dataset into training, validation, and test sets using an 80\%–10\%–10\% stratified split with a fixed random seed of 42 to ensure reproducibility. All experiments were run on a single NVIDIA T4 GPU with 16GB of memory. To account for variability, each configuration was executed five times, and we conducted a one‑way ANOVA on the resulting performance scores to assess the statistical significance of our findings. We have evaluated our models using accuracy, recall, precision and F1 scores. All our models have been implemented in PyTorch \cite{paszke2019pytorchimperativestylehighperformance} to improve ease of understanding. 

\section{Results}
In this section we first compare the two \textit{from–scratch} baselines (CNN\_\textit{base} and FNN\_\textit{base}), then present the impact of generalisation techniques (CNN\_\textit{gen}), followed by all transfer‑learning and multimodal variants.  

\paragraph{CNN and FNN}
\begin{table}[ht]
  \centering
  \resizebox{\columnwidth}{!}{%
  \begin{tabular}{lcccc}
    \toprule
    \textbf{Model} & \textbf{Accuracy} & \textbf{Precision} & \textbf{Recall} & \textbf{F1} \\
    \midrule
    $CNN_{base}$ & \textbf{0.3862} & \textbf{0.4204} & \textbf{0.4090} & \textbf{0.3910} \\
    $FNN_{base}$ & 0.3241 & 0.3028 & 0.3853 & 0.2851 \\
    \bottomrule
  \end{tabular}}
  \caption{Test performance of CNN\_base vs.\ FNN\_base.}
  \label{tab:cnn_fnn}
\end{table}
As Table~\ref{tab:cnn_fnn} illustrates, the \texttt{CNN\_base} model outperforms its fully connected counterpart --- raising accuracy from 32.4\% to 38.6\% and achieving higher precision and F1 score --- emphasising how convolutional layers’ spatial inductive biases yield richer feature representations than a parameter‑matched dense network.

\paragraph{Extending CNN}
\begin{table}[ht]
  \centering
  \resizebox{\columnwidth}{!}{%
  \begin{tabular}{lcccc}
    \toprule
    \textbf{Augmentation} & \textbf{Valid Acc.} & \textbf{Precision} & \textbf{Recall} & \textbf{F1} \\
    \midrule
    Vertical flip          & \textbf{0.5517} & 0.5934 & \textbf{0.5682} & \textbf{0.5474} \\
    Perspective transform  & 0.4828 & \textbf{0.6190} & 0.4773 & 0.4359 \\
    All combined           & 0.4483 & 0.1429 & 0.3333 & 0.2000 \\
    Random resized crop    & 0.4138 & 0.4151 & 0.4520 & 0.4167 \\
    Color jitter           & 0.3448 & 0.1667 & 0.1944 & 0.1693 \\
    Gaussian blur          & 0.3103 & 0.3089 & 0.3687 & 0.3105 \\
    Rotation (15°)         & 0.3103 & 0.2193 & 0.3182 & 0.2387 \\
    Baseline (no aug.)     & 0.2759 & 0.2103 & 0.3056 & 0.2389 \\
    Horizontal flip        & 0.2759 & 0.2500 & 0.3939 & 0.2222 \\
    Grayscale              & 0.2414 & 0.3016 & 0.3359 & 0.2118 \\
    \bottomrule
  \end{tabular}}
  \caption{Effect of different augmentations on CNN performance evaluations on the Validation set.}
  \label{tab:augmentations}
\end{table}
Table~\ref{tab:augmentations} summarizes validation performance across ten augmentation strategies. Simple geometric transforms, particularly vertical flips, nearly doubled baseline accuracy and substantially boosted recall, while perspective transforms yielded the highest precision with only a modest drop in overall accuracy. Random resized crops also provided consistent gains. In contrast, aggressive combinations or colour-based perturbations, such as colour jitter and grayscale, often degrade performance, indicating that excessive or semantically misleading distortions can hinder learning.

\begin{table}[ht]
  \centering
  \resizebox{\columnwidth}{!}{%
  \begin{tabular}{lcccc}
    \toprule
    \textbf{Model} & \textbf{Accuracy} & \textbf{Precision} & \textbf{Recall} & \textbf{F1} \\
    \midrule
    $CNN_{gen}$  & \textbf{0.4138} & \textbf{0.4561} & \textbf{0.4679} & \textbf{0.3941} \\
    $CNN_{base}$ & 0.3793 & 0.4000 & 0.4263 & 0.3635 \\
    \bottomrule
  \end{tabular}}
  \caption{Overall performance of the augmented CNN (\texttt{CNN\_gen}) vs.\ baseline.}
  \label{tab:cnn_gen_overall}
\end{table}

Informed by the augmentation study, we built \texttt{CNN\_gen} by applying vertical flips, perspective transforms and random resized crops, and by adding batch normalization and dropout within each convolutional block. As Table~\ref{tab:cnn_gen_overall} shows, this model outperforms \texttt{CNN\_base} across all key metrics on the test set --- boosting accuracy by over 3 points and delivering comparable improvements in precision, recall and F1 score. These results confirm that combining targeted augmentations with stronger regularization markedly enhances generalization.

\paragraph{Binary Classification}
\begin{figure}[t]
\includegraphics[width=1\columnwidth]{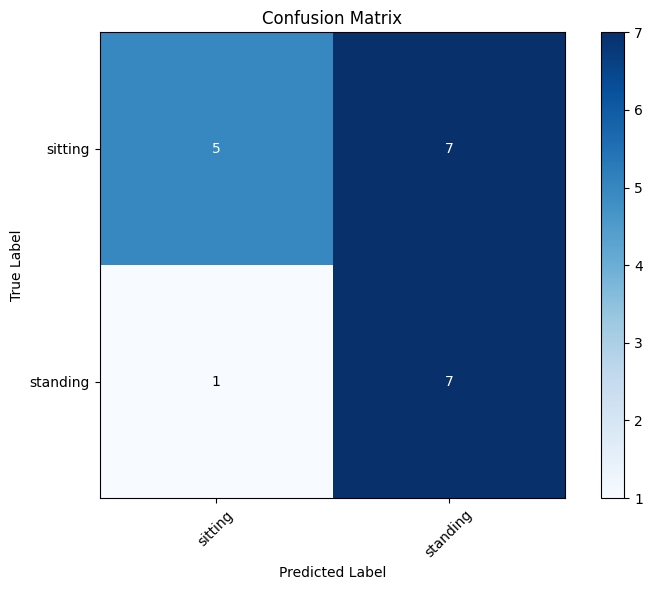}
\centering
\caption{Evaluation of ViT model trained on binary classes using transfer learning.}
\label{fig:binary_vit}
\end{figure}
\begin{table}[ht]
    \centering
    \resizebox{\columnwidth}{!}{%
    \begin{tabular}{lcccc}
        \toprule
        Model & Accuracy & Precision & Recall & F1 \\
        \midrule
        CLIP    & \textbf{0.65} & 0.6319 & \textbf{0.6250} & \textbf{0.6267} \\
        ViT     & 0.55 & \textbf{0.6333} & 0.6042 & 0.5396 \\
        Siglip2 & 0.50 & 0.5938 & 0.5625 & 0.4792 \\
        \bottomrule
    \end{tabular}}
    \caption{Performance metrics for binary classification models}
    \label{tab:model_metrics}
\end{table}

Table~\ref{tab:model_metrics} compares the binary classification performance when we classify the images into standing or sitting, three transfer‑learning models: CLIP, ViT, and SigLIP2. CLIP emerges clearly on top, leveraging its joint image–text training to capture class‑relevant cues that ViT and SigLIP2 miss. Although ViT matches CLIP’s ability to avoid false positives, it struggles to recall all positive instances --- often confusing subtle posture shifts --- while SigLIP2’s additional multilingual and dense pretraining appears to dilute its focus on our specific activities. Figure~\ref{fig:binary_vit}'s confusion matrix for ViT underscores these patterns, showing a notable fraction of sitting images incorrectly assigned to the standing class.

\paragraph{CLIP Embeddings}
\begin{table}[h]
    \centering
    \resizebox{\columnwidth}{!}{%
    \begin{tabular}{lcccc}
        \toprule
        \textbf{Model} &\textbf{Accuracy} & \textbf{Precision} & \textbf{Recall} & \textbf{F1} \\
        \midrule
        $CLIP_{EM}$   & \textbf{0.3793} &\textbf{ 0.4409} & \textbf{0.4103} & \textbf{0.3778} \\
        $CLIP_{CS}$ & 0.2759 & 0.0920 & 0.3333 & 0.1441 \\
        \bottomrule
    \end{tabular}}
    \caption{Performance metrics for different CLIP settings. $CLIP_{EM}$ denotes model trained on CLIP image embeddings, $CLIP_{CS}$ denotes model trained on CLIP image and text cosine similarities}
    \label{tab:model_performance}
\end{table}

\begin{figure*}[ht]
\includegraphics[width=1\textwidth]{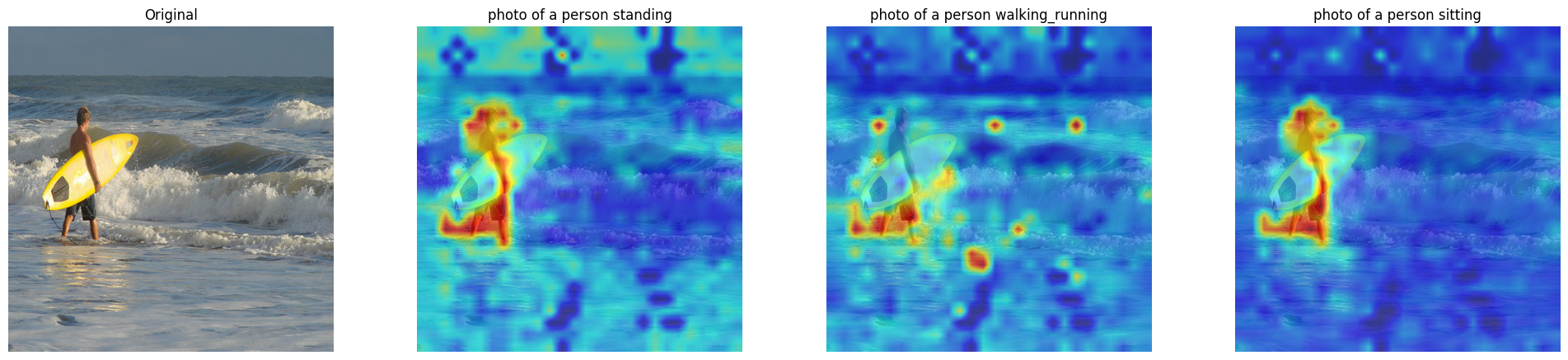}
\centering
\caption{Explainability of model}
\label{fig:legrad}
\end{figure*}

As Table~\ref{tab:model_performance} makes clear, the MLP built on raw CLIP embeddings outperforms its cosine‐similarity‐only counterpart across the board --- lifting accuracy from 27.6\% to 37.9\% and yielding correspondingly higher precision and F1. 

These results indicate that the high‑dimensional CLIP image vectors contain rich discriminative information that a shallow MLP can effectively exploit, whereas relying only on the \(C\)-dimensional similarity scores compresses the features too aggressively and hampers class separation.  In particular, the steep drop in precision for \(CLIP_{CS}\) suggests it produces many false positives when forced to decide based on cosine relationships alone.

\paragraph{Multiclass Classification}
\begin{table}[ht]
  \centering
  \resizebox{\columnwidth}{!}{%
  \begin{tabular}{lcccc}
    \toprule
    \textbf{Model}      & \textbf{Test Accuracy}   & \textbf{Precision}      & \textbf{Recall}         & \textbf{F1 Score}       \\
    \midrule
    $CLIP_{IC}$                & $0.759_\pm{}_{0.000}$     & $0.743_\pm{}_{0.000}$     & $0.740_\pm{}_{0.000}$     & $0.737_\pm{}_{0.000}$     \\
    Siglip2             & $0.634_\pm{}_{0.028}$     & $0.635_\pm{}_{0.025}$     & $0.648_\pm{}_{0.021}$     & $0.636_\pm{}_{0.025}$     \\
    ViT                 & $0.462_\pm{}_{0.017}$     & $0.462_\pm{}_{0.029}$     & $0.437_\pm{}_{0.035}$     & $0.434_\pm{}_{0.026}$     \\
    $CNN_{gen}$          & $0.386_\pm{}_{0.114}$     & $0.420_\pm{}_{0.127}$     & $0.409_\pm{}_{0.111}$     & $0.391_\pm{}_{0.116}$     \\
    $CNN_{base}$            & $0.379_\pm{}_{0.079}$     & $0.441_\pm{}_{0.131}$     & $0.417_\pm{}_{0.088}$     & $0.369_\pm{}_{0.078}$     \\
    $CLIP_{CS}$        & $0.359_\pm{}_{0.052}$     & $0.120_\pm{}_{0.017}$     & $0.333_\pm{}_{0.000}$     & $0.175_\pm{}_{0.018}$     \\
    $CLIP_{EM}$           & $0.331_\pm{}_{0.094}$     & $0.356_\pm{}_{0.112}$     & $0.348_\pm{}_{0.096}$     & $0.301_\pm{}_{0.073}$     \\
    $FNN_{base}$           & $0.324_\pm{}_{0.052}$     & $0.303_\pm{}_{0.074}$     & $0.385_\pm{}_{0.069}$     & $0.285_\pm{}_{0.047}$     \\
    \bottomrule
  \end{tabular}}
  \caption{Final leaderboard. $CLIP_{IM}$ denotes a model trained on CLIP image embeddings, $CLIP_{CS}$ denotes a model trained on CLIP image and text cosine similarities. $CLIP_{IC}$ denotes the CLIP model finetuned for classification.}
  \label{tab:final_leaderboard}
\end{table}

Table~\ref{tab:final_leaderboard} presents the full multiclass comparison. The fine‑tuned \texttt{CLIP\_IC} model leads by a wide margin, achieving roughly 76\% accuracy and similarly high precision, recall, and F1. It outperforms the next best, SigLIP2, and leaves ViT and our convolutional baselines (\texttt{CNN\_gen}, \texttt{CNN\_base}) trailing well behind. The shallow embedding‑based classifiers ($CLIP_{EM}$, $CLIP_{CS}$) occupy the middle ground, while the fully connected \texttt{FNN\_base} sits at the bottom. A one‐way ANOVA on accuracy across the seven models confirms that these differences are statistically significant (F=23.4562, p<0.001).  A paired t‐test between the two from‐scratch baselines (\texttt{CNN\_base} vs.\ \texttt{FNN\_base}) yields t=1.4505, p=0.2205, indicating no significant difference between them.  Together, these results highlight the clear advantage of large‐scale, contrastive pretraining (as in CLIP and SigLIP2) over both vanilla convolutional and dense architectures, and the particular strength of CLIP when fine‐tuned for our multiclass classification problem.  

\section{Discussion}

\paragraph{Explainablity}
To better understand how our best model that is CLIP distinguishes between visually similar human activities, we apply LeGrad \cite{bousselham2024legrad}, an explainability method tailored for transformer‐based vision models.  LeGrad computes the gradient of the model’s output logits with respect to each attention map across all ViT layers, then aggregates these signals --- combining both intermediate and final token activations --- into a single saliency map.  In Figure~\ref{fig:legrad}, we show the original image (a) alongside the LeGrad maps for model’s predicted classes “standing” (b), “walking\_running” (c), and “sitting” (d).
These visualisations reveal two key challenges posed by our dataset.  First, the fine-grained differences between standing, walking, and sitting settings result in overlapping attention regions, which can confuse the model’s decision boundary.  Second, the heterogeneous objects in the image introduce noise into the attention gradients, making it difficult for even a powerful transformer-based encoder to focus exclusively on the human subject.  Together, these factors help explain why our highest‐accuracy models still struggle to exceed 80\% on these classes and underscore the need for more targeted spatio‐temporal features or refined attention mechanisms.

\paragraph{Error Analysis}
Figure~\ref{fig:error_analysis} illustrates representative failure cases of our models on “standing,” “walking\_running,” and “sitting.” We observe that small or partially occluded people are often mistaken for static poses by CNN\_base and FNN\_base, while low‐amplitude motions (e.g.\ slow walking) confuse ViT and Siglip2, which over‑rely on per‐frame posture cues. Dynamic backgrounds (e.g.\ moving vehicles or flags) occasionally dominate Siglip2’s embeddings, leading to “standing” predictions, whereas CLIP’s multimodal pretraining improves robustness but still misclassifies very low‐resolution actors. Finally, borderline poses (e.g.\ slight weight shifts) lie near the decision boundary for all models. These systematic errors underscore the need for richer spatio-temporal features and stronger human-focused attention mechanisms to further enhance performance.

\section{Conclusion}

Our results demonstrate that while traditional CNN benefit from well‑chosen augmentations and regularisation, the most dramatic gains arise from contrastive image–text pretraining, which yields far more discriminative feature spaces. Inspection of model visualisations revealed that attention sometimes drifts to background elements, and the hardest errors occur when poses are partially obscured or lie near decision boundaries.

\noindent\textbf{Total Words}: 2092

% Entries for the entire Anthology, followed by custom entries
\bibliography{custom}

\appendix

\section{Appendix}
\label{sec:appendix}

\begin{figure*}[ht]
\includegraphics[width=1\textwidth]{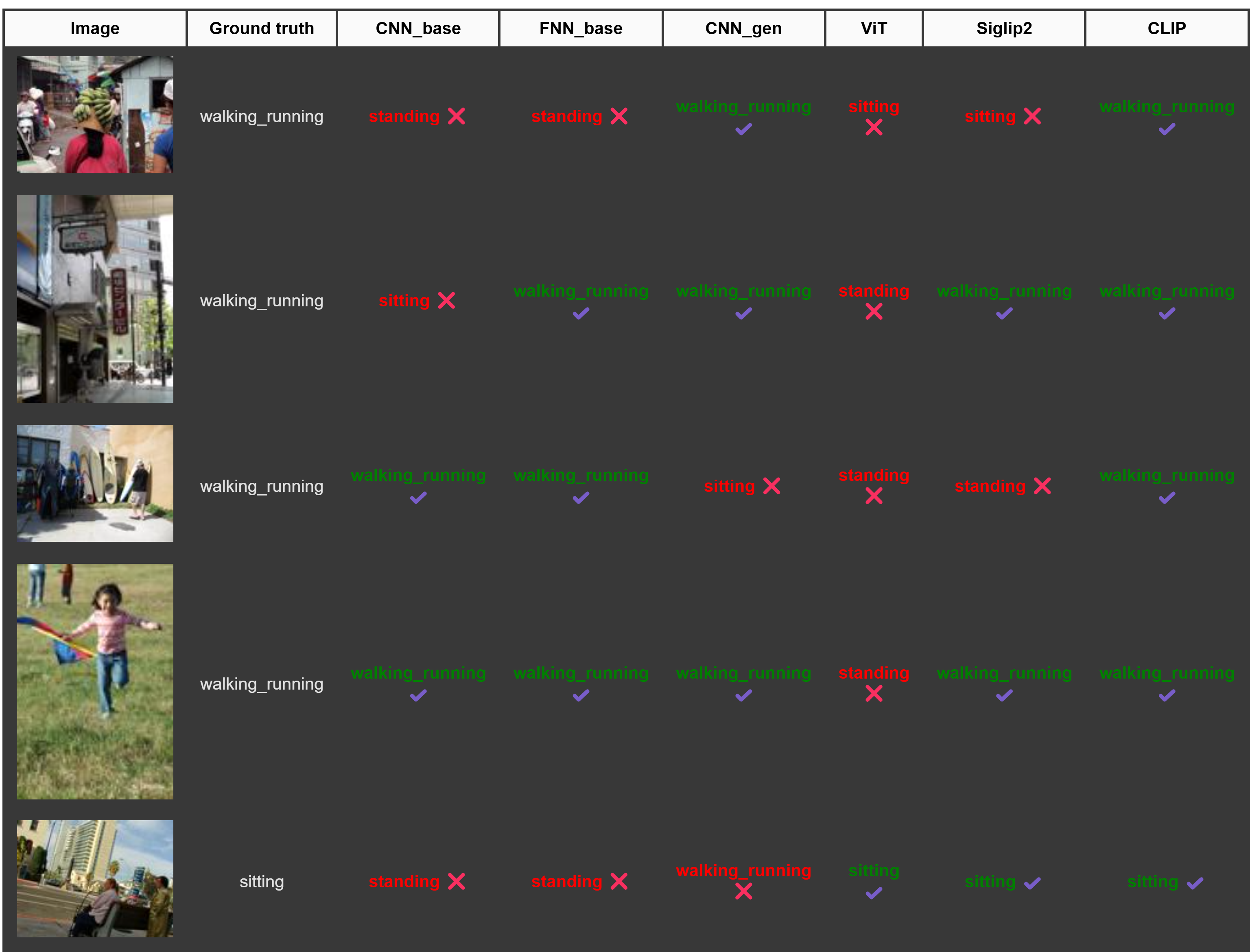}
\centering
\caption{Error analysis of the models}
\label{fig:error_analysis}
\end{figure*}

\end{document}